%% file: main.tex
\documentclass[sigconf]{acmart}
\AtBeginDocument{%
  \providecommand\BibTeX{{%
    \normalfont B\kern-0.5em{\scshape i\kern-0.25em b}\kern-0.8em\TeX}}}

\input{HierTCN-KDD22/head}
\input{HierTCN-KDD22/redefine}

\input{HierTCN-KDD22/abstract}

\begin{document}
\input{HierTCN-KDD22/title}
\maketitle

\input{HierTCN-KDD22/introduction}
\input{HierTCN-KDD22/dataset}
\input{HierTCN-KDD22/related_work}
\input{HierTCN-KDD22/method}
\input{HierTCN-KDD22/experiment}

\input{HierTCN-KDD22/conclusion}

\bibliographystyle{acm}
\bibliography{main.bib}

\clearpage
\input{HierTCN-KDD22/supplementary}

\end{document}

%% file: HierTCN-KDD22/head.tex
\copyrightyear{2022}
\acmYear{2022}
\setcopyright{rightsretained}
 
\acmConference[KDD '22]{Proceedings of the 28th ACM SIGKDD Conference on Knowledge Discovery and Data Mining}{August 14--18, 2022}{Washington, DC, USA} \acmBooktitle{Proceedings of the 28th ACM SIGKDD Conference on Knowledge Discovery and Data Mining (KDD '22), August 14--18, 2022, Washington, DC, USA} \acmDOI{10.1145/3534678.3539056} \acmISBN{978-1-4503-9385-0/22/08}
 
\usepackage{etoolbox}
\makeatletter
\patchcmd{\maketitle}{\@copyrightpermission}{
  \begin{minipage}{0.3\columnwidth}
     \href{https://creativecommons.org/licenses/by-nc-sa/4.0/}{\includegraphics[width=0.90\textwidth]{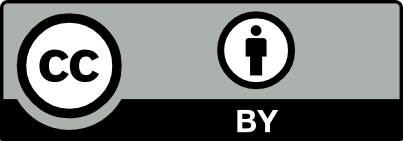}}
  \end{minipage}\hfill
  \begin{minipage}{0.7\columnwidth}
     \href{https://creativecommons.org/licenses/by-nc-sa/4.0/}{This work is licensed under a Creative Commons Attribution-NonCommercial-ShareAlike International 4.0 License.}
  \end{minipage}
 
  \vspace{5pt}
}{}{}
 
\makeatother



%% file: HierTCN-KDD22/redefine.tex
\usepackage[linesnumbered,ruled,vlined,algo2e]{algorithm2e}
\usepackage{multirow}
\usepackage{tabularx}
\usepackage{graphicx}
\usepackage{float}
\usepackage{subfigure}
\usepackage{array}
\usepackage{amsfonts}
\usepackage{epsfig}
\usepackage{graphicx}
\usepackage{amsmath}
\usepackage{booktabs}
\usepackage{appendix}
\usepackage{bbm}
\usepackage{stfloats}
\usepackage{flushend}

\usepackage{amsthm}
\usepackage{mathtools}
\usepackage{commath}

\newcommand{\PreserveBackslash}[1]{\let\temp=\\#1\let\\=\temp}
\newcolumntype{C}[1]{>{\PreserveBackslash\centering}p{#1}}
\newcolumntype{R}[1]{>{\PreserveBackslash\raggedleft}p{#1}}
\newcolumntype{L}[1]{>{\PreserveBackslash\raggedright}p{#1}}

\usepackage{enumitem}
\setlist[itemize]{leftmargin=*}

\newcommand{\ours}{HiPAL\xspace}
\newcommand{\our}{HiPAL}

\newcommand{\simtcn}{CausalNet\xspace}
\newcommand{\restcn}{ResTCN\xspace}

\newcommand{\higru}{HierGRU\xspace}
\newcommand{\hifcn}{\our-{f}\xspace}
\newcommand{\hisimtcn}{\our-{c}\xspace}
\newcommand{\hirestcn}{\our-{r}\xspace}

\newcommand{\action}{\mathbf{e}}
\newcommand{\actionEmb}{\mathbf{a}}
\newcommand{\intervEmb}{\mathbf{b}}
\newcommand{\stampEmb}{\mathbf{c}}
\newcommand{\emb}{\mathbf{g}}
\newcommand{\shift}{\mathbf{h}}
\newcommand{\mon}{\mathbf{v}}
\newcommand{\subs}{_i^{(k)}}

%% file: HierTCN-KDD22/abstract.tex
\begin{abstract}
Burnout is a significant public health concern affecting nearly half of the healthcare workforce. This paper presents the first end-to-end deep learning framework for predicting physician burnout based on electronic health record (EHR) activity logs, digital traces of physician work activities that are available in any EHR system. In contrast to prior approaches that exclusively relied on surveys for burnout measurement, our framework directly learns deep representations of physician behaviors from large-scale clinician activity logs to predict burnout. We propose the \underline{Hi}erarchical burnout \underline{P}rediction based on \underline{A}ctivity \underline{L}ogs (\ours), featuring a pre-trained time-dependent activity embedding mechanism tailored for activity logs and a hierarchical predictive model, which mirrors the natural hierarchical structure of clinician activity logs and captures physicians' evolving burnout risk at both short-term and long-term levels. 
To utilize the large amount of unlabeled activity logs, we propose a semi-supervised framework that learns to transfer knowledge extracted from unlabeled clinician activities to the \our-based prediction model. 
The experiment on over 15 million clinician activity logs collected from the EHR at a large academic medical center demonstrates the advantages of our proposed framework in predictive performance of physician burnout and training efficiency over state-of-the-art approaches.

\end{abstract}

\ccsdesc[500]{Applied computing~Health informatics}
\ccsdesc[500]{Computing methodologies~Temporal reasoning}
\ccsdesc[500]{Computing methodologies~Neural networks}

\keywords{EHR activity logs, physician wellness, deep sequence learning}

%% file: HierTCN-KDD22/title.tex
\title[HiPAL: A Deep Framework for Physician Burnout Prediction Using Activity Logs]{HiPAL: A Deep Framework for Physician Burnout Prediction Using Activity Logs in Electronic Health Records}


\author{Hanyang Liu}
\affiliation{%
  \institution{McKelvey School of Engineering \\ Washington University in St. Louis}
  \streetaddress{1 Brooking Drive}
  \city{St. Louis}
  \country{United States}
  \postcode{63130}
}
\email{hanyang.liu@wustl.edu}

\author{Sunny S. Lou}
\affiliation{%
  \institution{School of Medicine \\ Washington University in St. Louis}
  \city{St. Louis}
  \country{United States}
  \postcode{63130}
}
\email{slou@wustl.edu}

\author{Benjamin C. Warner}
\affiliation{%
  \institution{McKelvey School of Engineering \\ Washington University in St. Louis}
  \streetaddress{1 Brooking Drive}
  \city{St. Louis}
  \country{United States}
}
\email{b.c.warner@wustl.edu}

\author{Derek R. Harford}
\affiliation{%
  \institution{School of Medicine \\ Washington University in St. Louis}
  \city{St. Louis}
  \country{United States}
}
\email{derek.harford@wustl.edu}

\author{Thomas Kannampallil}
\affiliation{%
  \institution{School of Medicine \\ Washington University in St. Louis}
  \city{St. Louis}
  \country{United States}
}
\email{thomas.k@wustl.edu}

\author{Chenyang Lu}
\authornote{Corresponding author.}
\affiliation{%
  \institution{McKelvey School of Engineering \\ Washington University in St. Louis}
  \streetaddress{1 Brooking Drive}
  \city{St. Louis}
  \country{United States}
}
\email{lu@wustl.edu}

\renewcommand{\shortauthors}{Hanyang Liu et al.}

%% file: HierTCN-KDD22/introduction.tex
\section{Introduction}

Burnout is a state of mental exhaustion caused by one’s professional life. 
It contributes to poor physical and emotional health, and may lead to alcohol abuse and suicidal ideation \cite{hu2019discrimination}.
Physician burnout is widespread in healthcare settings and affects nearly 50\% of physicians and health workers. It is associated with negative consequences for physician health, their retention, and the patients under their care \cite{national2019taking}. The recent COVID-19 pandemic has further highlighted the negative impact of physician burnout \cite{prasad2021prevalence}.  In essence, burnout is a considerable public health concern, and effective tools for monitoring and predicting physician burnout are desperately needed \cite{national2019taking}.

One of the key contributors to burnout is physician behavior (e.g., workload, workflow) \cite{lou2021temporal}. 
The availability of electronic health record (EHR) activity logs, the digital footprint of physicians' EHR-based activities, has enabled studies for measuring physician workload \cite{sinsky2020metrics,lou2021temporal}, workflow \cite{chen2021mining}, and cognitive burden \cite{lou2022effect}, illustrating the considerable potential of EHR activity logs to capture physician behaviors at the microscale. 
More recently, such activity logs have been used to predict burnout using off-the-shelf machine learning models \cite{lou2021predicting,escribe2022understanding}. However, as these models are unable to directly process unstructured data, they rely exclusively on feature engineering, using hand-crafted summary statistics of activity logs as features. Developing such models, hence, requires considerable domain knowledge in medicine and cognitive psychology to obtain clinically meaningful measures, and these statistical features are often less effective in capturing the complicated dynamics and temporality of activities.

An ideal burnout prediction framework should be end-to-end, i.e., able to efficiently learn deep representations of physician behavior directly from raw activity logs. This enables the potential for real-time phenotyping of burnout that is of high clinical value in facilitating early intervention and mitigation for the affected physician. 
There are two major challenges in building such a framework. 
The \textit{first challenge} is to extract useful knowledge from unstructured raw log files, which track over 1,900 types of EHR-based actions (activities associated with viewing reports, notes, laboratory tests, managing orders, and documenting patient care activities).  A predictive framework must be able to encode both these activities and associated timestamps and capture the underlying dynamics and temporality, which builds up to high-level representations. In other words, an effective data encoding mechanism tailored for activity logs is key for any deep sequence model. 

The \textit{second challenge} for training a deep predictive model for burnout is the large scale of activity logs (i.e., long data sequences) and limited number of labels (i.e., relatively few burnout survey responses). 
In order to measure physician behavior and predict burnout at a per-month basis, the sequence model must be able to efficiently process sequences with large-scale and highly variant length from a few hundred to over 90,000 events. On the other hand, due to the high cost and uncertainty of survey collection, only half of the activity logs recorded are labeled with burnout outcomes. These require the sequence model to have not only a long-term memory (i.e., wide range of receptive field \cite{bai2018empirical}) but also a relatively small model complexity (i.e., number of parameters) to prevent overfitting.
However, many popular sequence models based on recurrent neural networks (RNN) or 1D Transformer \cite{zhou2021informer,beltagy2020longformer} are not suitable for raw activity logs of this large scale due to high time or memory cost.

Apart from addressing the above-mentioned two challenges, it would be useful for a predictive model to further capture and utilize the hierarchical structure naturally embedded in physician activity logs (see Figure \ref{fig:teaser}). 
Physicians' work life is intrinsically hierarchical.
They interact with the EHR system in sessions -- clusters of activities -- with various lengths that are embedded within a single work shift, and then work multiple shifts per month. Intuitively, the temporal congregation of clinical activities may contain useful information associated with burnout, i.e., the same total workload spread evenly over a week likely has a different effect on wellness than more intense work spread over only two days.
However, the single-level sequence models are unable to unobtrusively recover the hierarchical structure or the multi-scale temporality of data.
And none of the recently proposed hierarchical sequence models such as \cite{zhao2017hierarchical,yang2016hierarchical,taylor2017personalized} are designed for burnout prediction or similar problems, nor are they efficient enough in processing sequences at this large scale. 


To addresses these challenges, we propose the \underline{Hi}erarchical burn-out \underline{P}rediction based on \underline{A}ctivity \underline{L}ogs (\ours), a deep learning framework from representation learning to hierarchical sequence modeling as a comprehensive solution to the burnout prediction problem.
To the best of our knowledge, this is the first end-to-end approach that directly uses raw EHR activity logs for burnout prediction. \ours features the following key components:
\vspace{-0.5em}
\begin{itemize}
    \item 
    A pre-trained time-dependent activity embedding mechanism tailored for EHR activity logs that is designed to encode both the dynamics and temporality of activities.
    The encoded joint representations build up and shape the bottom-up activity measures at multiple levels as the basis for burnout prediction.
    
    \item We propose a novel and generalizable hierarchical deep architecture that can build upon any existing efficient sequence model for burnout prediction. A low-level sequence encoder aggregates short-term activities into deep daily activity measures, whereas the high-level RNN-based encoder further cumulatively arranges all activity measures into deep monthly activity measures and captures the long-term temporality of physician behaviors. The hierarchical architecture with shared low-level encoder enables \ours to maintain long memory with multi-scale temporality without increasing the model complexity. 
    
    \item To utilize the large amount of unlabeled activity logs, we extend \ours to a semi-supervised framework (Semi-\ours). For this, we pre-train the low-level encoder and transfer knowledge to the predictive model with an unsupervised sequence autoencoder that learns to reconstruct the action-time sequences. 
    
    \item The experiment on a real-world dataset of over 15 million activity logs from 88 physicians over a 6-month period shows improved performance and high computational efficiency of \ours and Semi-\ours in predicting burnout. 
    
\end{itemize}

%% file: HierTCN-KDD22/dataset.tex
\section{Burnout Prediction Problem}

\begin{figure}
    \centering
    \includegraphics[width=0.98\linewidth]{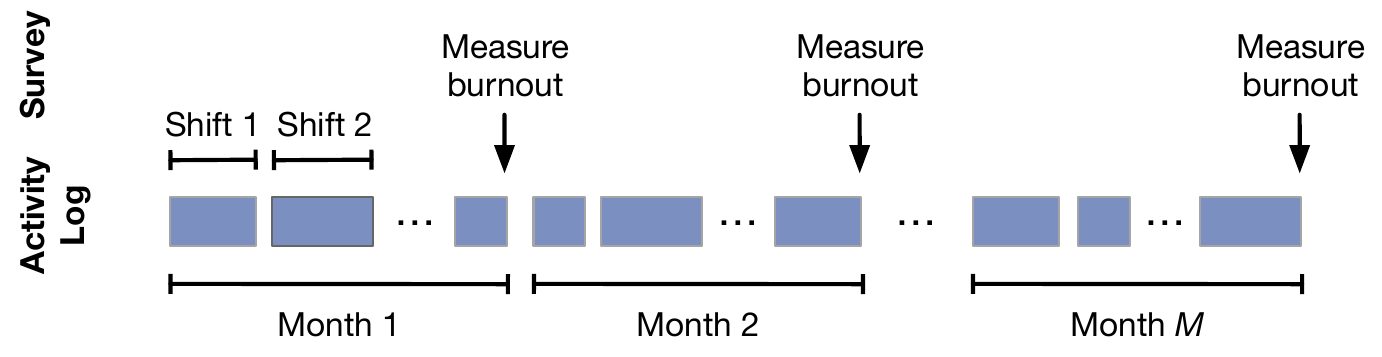}
    \caption{Illustration of EHR activity logs with shift-month hierarchical structure and monthly surveys.}
    \label{fig:teaser}
\end{figure}

\subsection{Data}
The data used in this work were collected from 88 intern and resident physicians in Internal Medicine, Pediatrics and Anesthesiology at the Washington University School of Medicine, BJC HealthCare
and St Louis Children’s Hospital
During the data collection phase from September 2020 through April 2021, all participants consented (IRB\# 202004260) to provide access to their EHR-based \textit{activity logs} and to complete \textit{surveys} every month. 


\begin{figure}[t]
    \centering
    \includegraphics[width=0.99\linewidth]{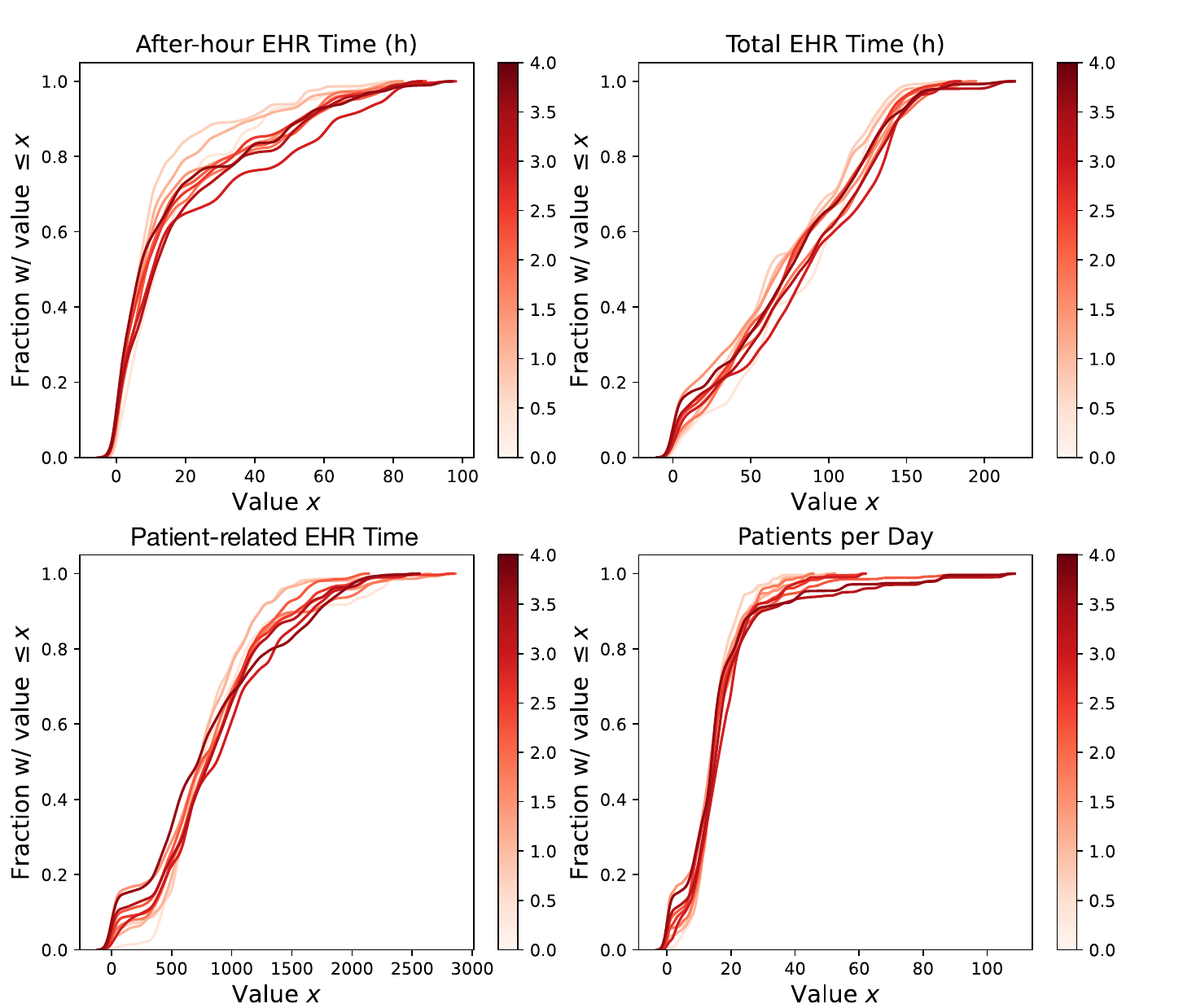}
    \caption{CDFs of hand-crafted clinical activity measures of EHR activity logs on a monthly basis grouped by various burnout score ranges from low to high.}
    \label{fig:ccdf}
\end{figure}

\subsubsection{EHR Activity Logs} 
EHR activity logs are traces of a clinician's interactions with the EHR. 
These log files record all activities performed on an EHR system including the time, patient, activity, and user responsible for each data access event, and are a comprehensive record of a clinician’s work activities.
In our dataset, there were 1,961 distinct types of clinical activities such as review of patient data, note writing, order placement and review of clinical inbox messages. Each activity log action (i.e., clinical activity) of the user (i.e., physician participant) was recorded by the EHR system with a timestamp specifying the start time of each action. All activity log actions are recorded passively in system background without interfering with the user's normal clinical workflow. In total, over 15 million activity logs across 6 to 8 months were collected in the dataset (on average over 20,000 logs per month per participant).

\subsubsection{Burnout Surveys} 
Intern and resident physicians included in this study rotated between different clinical assignments (e.g., Internal Medicine, Pediatrics, and Anesthesiology) every 4 weeks. Surveys were designed to evaluate each participant's recent wellness status and were sent to each participant at 4-week intervals, timed to coincide with the end of each rotation. Each participant is asked to complete 6 surveys. The monthly surveys were used to evaluate the participant's burnout status using the Stanford Professional Fulfillment Index (PFI) \cite{trockel2018brief} based on workload exhaustion and depersonalization, with scores ranging from 0 to 4. We follow the previous work \cite{lou2021predicting} to define burnout as the PFI score being greater than or equal to 1.33. 

\subsubsection{Work Shift}
A physician's shiftwork may occur during the day or night with the start and end time varying over rotations and individuals. 
The work within a shift is in general continuous with relatively short intervals between activities. 
Typically a physician has one shift per day.
The work of a monthly rotation is naturally segmented into separate temporal clusters by shifts (see Figure \ref{fig:teaser}).
We follow \cite{dziorny2019automatic} to automatically segment the activity logs based on the lengths of intervals of recorded time stamp.

\subsubsection{Relationship Between Activity and Burnout}
As part of preliminary data analysis, we assessed the correlation between several hand-crafted activity measures (e.g., workload) and burnout scores. We selected several basic summary statistics (e.g., time spent on EHR) for assessment. 
The cumulative distribution function (CDF) of each measurement is displayed in Figure \ref{fig:ccdf}. Each group of physicians with different burnout severity range on a monthly basis are colored progressively (darker colors mean more severe burnout).
In general, there was an association between physician workload and risk of burnout. 
Meanwhile, some activity measures (e.g., patient related EHR time) appeared progressive and had a complicated associative pattern.
Hence, there existed a certain level of predictive information in the EHR activity logs that can be extracted via designed activity measures and then used for burnout prediction as in our prior work \cite{lou2021predicting}.
Nevertheless, the design and selection of clinically meaningful activity measures required considerable domain knowledge. These delicately designed measures (usually with summary statistics as in Figure \ref{fig:ccdf}) showed limited effectiveness in capturing the complicated predictive patterns \cite{lou2021predicting}. An end-to-end deep learning framework is therefore needed to learn complex representations of physician behaviors from raw activity logs.

\subsection{Problem Formulation}
Formally, in longitudinal EHR activity log data, the activities of the $m$-th month for the $n$-th clinician can be represented as $\mathbf{P}_{m, n} = \{ \mathcal{V}^{(k)} \}_{k=1}^{T_{m,n}}$, where $T_{m,n}$ is the number of shifts in the $m$-th month for clinician $n$. The whole activity log dataset can be written as $\{\{\mathbf{P}_{m, n}\}^{M_n}_{m=1}\}^{N}_{n=1}$, where $M_n$ is the total number of survey months of data collected from clinician $n$ and $N$ is the total number of clinicians.
Here the indices of clinicians $n$ and survey months $m$ are omitted for simplicity.
The work at the $k$-th shift for a clinician can be represented as a sequence of actions and their time stamps $\mathcal{V}^{(k)} = [ (\action_{1}^{(k)}, t_{1}^{(k)}), (\action_{2}^{(k)}, t_{2}^{(k)}), ..., (\action_{N_k}^{(k)}, t_{N_k}^{(k)}) ]$, where $\action_i^{(k)}\in \{0, 1\}^{|\mathcal{A}|}$ is the one-hot vector denoting the $i$-th action at time $t_i$, $N_k$ is the number of actions in the $k$-th shift, $\mathcal{A}$ is the set of clinician actions, and $|\mathcal{A}|$ is the number of unique actions among all clinicians. The goal is to use the activity logs, $\mathbf{P}_{m, n}$, to predict the binary label $y_{m, n} \in \{0, 1\}$ that denotes the wellness status of clinician $n$ in the $m$-th month, by learning a predictive model $f: f(\mathbf{P}_{m, n}) \rightarrow y_{m, n}$. 


%% file: HierTCN-KDD22/related_work.tex
\section{Related Work}
\textbf{Burnout Prediction.}
Early studies mainly focused on risk factors associated with burnout using self-reported surveys and have highlighted the link between perceived workload and burnout \cite{shanafelt2009burnout,shanafelt2012burnout}. Machine learning models such as k-means and linear regression were used to predict the level of burnout from survey responses \cite{bauernhofer2018subtypes,lee2020app,batata2018mixed}.
Because these approaches relied purely on self-reported survey responses as input data, they are subject to inaccuracy in workload measurement and are unable to provide unobtrusive burnout prediction \cite{kannampallil2021conceptual}.
Recently, the quantification of clinician workload based on EHR activity logs has enabled studies to track EHR-based activity measures \cite{baxter2021measures,sinsky2020metrics},
and to apply off-the-shelf machine learning to predict burnout based on delicately design summary statistics of clinical activities as features \cite{lou2021predicting,escribe2022understanding}.

\vspace{0.5em}
\noindent\textbf{Sequence Models.}
The directional nature of Recurrent Neural Networks (RNN) and its popular variants LSTM and GRU \cite{chung2014empirical} has made them the default choices for modeling sequential data such as natural language, time series, and event sequences. However, RNN-based models are slow and difficult to train on large-scale sequential data due to its step-by-step recurrent operation and potential issues with long-term gradient backpropagation \cite{you2019hierarchical}. 
Recently, 1D Transformer variants \cite{zhou2021informer,beltagy2020longformer} have applied multi-head self-attention \cite{vaswani2017attention} to time series and natural language problems and seek to model long dependencies, but they still have significant time and space complexity, higher than RNNs.
In contrast, convolutional neural networks (CNN) based models are more efficient in modeling long sequences due to their parallelism. 
Fully Convolutional Networks (FCN) and Residual Convolutional Networks (ResNet) \cite{wang2017time} and have demonstrated superior performance in time series classification on over 40 benchmarks. Meanwhile, 1D CNNs with dilated causal convolutions have shown promise in efficiently modeling long sequential data such as audio \cite{oord2016wavenet}. This idea has been further extended and developed as a class of long-memory sequence models, Temporal Convolutional Networks (TCN), to model large-scale sequential data such as videos and discrete events \cite{lea2017temporal,bai2018empirical,dai2019self}. In ResTCN \cite{bai2018empirical}, multiple layers of dilated causal convolutions are stacked together to form a block combined with residual connections between blocks in order to build deeper networks efficiently. 

\vspace{0.5em}
\noindent\textbf{Hierarchical Sequence Models.}
Hierarchical models have shown promise in capturing the hierarchical structure of data \cite{yang2016hierarchical,zhao2017hierarchical, lin2015hierarchical} or obtaining multi-scale temporal patterns \cite{you2019hierarchical,quadrana2017personalizing} from long sequences such as document \cite{yang2016hierarchical}, videos \cite{zhao2017hierarchical}, online user record \cite{you2019hierarchical,quadrana2017personalizing,song2019hierarchical}, and sensor data \cite{phan2019seqsleepnet,du2015hierarchical}. 
However, even with a multi-level architecture, RNN-based hierarchical models \cite{yang2016hierarchical,lin2015hierarchical} would still struggle with efficiency in processing long sequences.
Recently, \cite{you2019hierarchical} proposed to build a hierarchical model on top of TCN for efficiently
modeling multi-scale user interest for a recommendation system, in which the TCN is used as the decoder for sequence generation conditioned on a high-level RNN for long-term dependencies. Despite the similar motivation, \cite{you2019hierarchical} is unsuitable for classification due to the sequence-to-sequence architecture designed for recommendation systems. 
In general, none of these approaches were designed for burnout prediction or similar problems, nor tailored for EHR activity logs with unique data modality and hierarchical structure significantly distinct from all the above applications.

%% file: HierTCN-KDD22/method.tex
\begin{figure*}[t]
    \centering
    \includegraphics[width=0.99\linewidth]{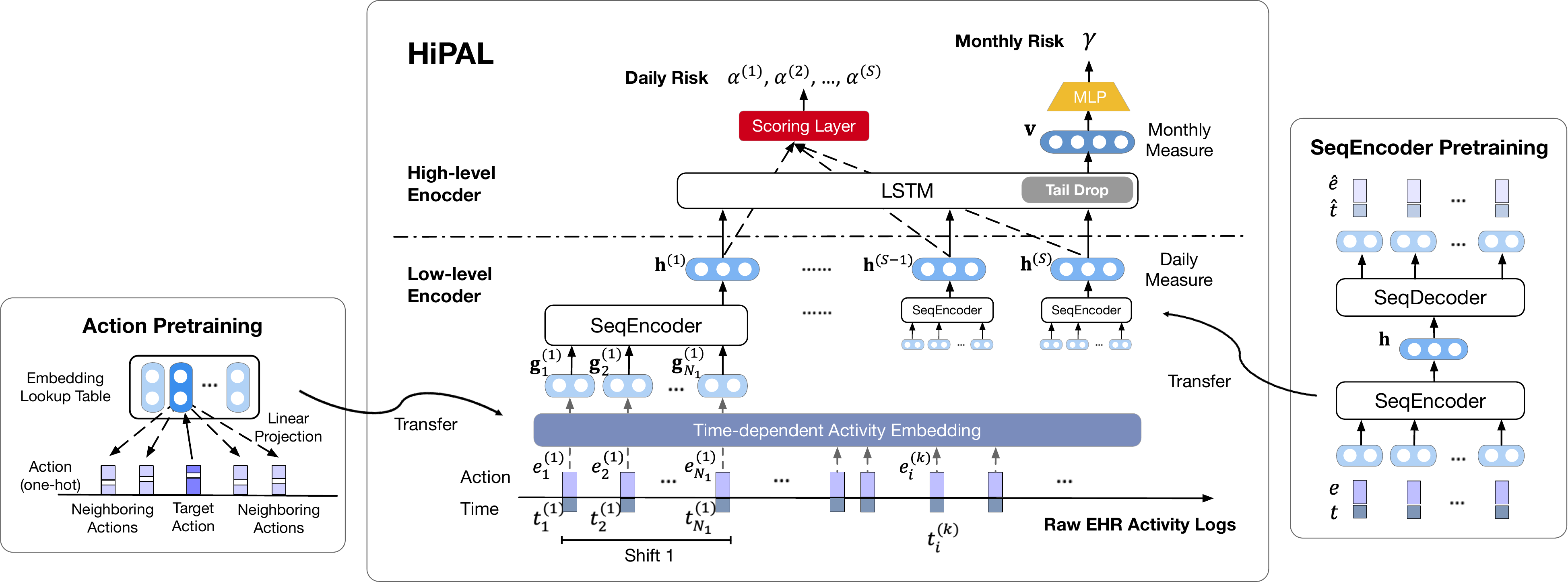}
    \caption{Overview of \ours physician burnout prediction framework.}
    \label{fig:framework}
\end{figure*}

\section{The \ours Framework}
Figure \ref{fig:framework} shows the overview of our \ours framework, featuring a pre-trained time-dependent activity embedding mechanism tailored for activity logs, a hierarchical predictive model that models physician behaviors in multiple temporal levels, and the semi-supervised framework that transfers knowledge from unlabeled data. Our \ours framework is generalizable to building upon any convolution-based sequence model as the base model.

\subsection{Time-dependent Activity Embedding}
\label{sec:embedding}

Different from other sequential data such as natural language and time series, activity logs consist of sequences of clinical actions and associated timestamps.
The physician workflow and workload information associated with the risk of burnout are preserved in the dynamics and temporality within these ordered action-time pairs.
We design a specific action-time joint embedding method to extract these patterns for prediction.

\vspace{-0.5em}
\subsubsection{Encoding Actions}
The one-hot action code $\action_i^{(k)}\in \mathbb{B}^{|\mathcal{A}|}$ is linearly mapped to a lower-dimensional embedding, 
$\mathbf{a}\subs = \mathbf{W}_a\action\subs$, where $ \mathbf{a}\subs \in \mathbb{R}^d$.
Here $\mathbf{W}_a \in \mathbb{R}^{d\times |\mathcal{A}|}$ is the embedding matrix where each column vector of it represents the lower-dimensional embedding of each unique action.

\textbf{Context-aware Pre-training. }
The representation of action $\mathbf{b}$ should be able to encode the contextual information, i.e., similar embedding $\mathbf{b}$ should represent actions with similar context. 
We propose to pre-train the embedding matrix $\mathbf{W}_a$ in an unsupervised fashion inspired by Word2Vec \cite{mikolov2013efficient} word embedding that has been widely used in many natural language processing tasks. Here we adopt skip-gram \cite{mikolov2013efficient} for embedding pre-training. The embedding of the $i$-th action $\action_i$ is linearly projected to predict actions $\{ \action_{i-L}, ..., \action_{i-1}, \action_{i+1}, ..., \action_{i+L} \}$. We set $L=5$ in this paper.

\vspace{-0.5em}
\subsubsection{Encoding Time Intervals}
The time information can be fully represented as the time intervals between two adjacent actions.
We first obtain the time interval sequence from the original time stamps $[t_1^{(k)}, t_2^{(k)}, ..., t_{N_k}^{(k)}]$ and then map the log-transformed time interval to a vector of $d$ dimensions, $\intervEmb\subs \in \mathbb{R}^{d}$
\begin{equation}
\label{eq:interv}
    \intervEmb\subs = \text{tanh}\left(\mathbf{W}_b\cdot \log\left(t\subs - t_{i - 1}^{(k)}\right) + \mathbf{d}_b \right)
\end{equation}
where $\mathbf{W}_b$ and $\mathbf{d}_b$ are the weight and bias variables for the time interval embedding. 
The time intervals with long-tail distribution are log-transformed to obtain embedding with more stationary dynamics.
Note that for the first action, $\intervEmb_{1}^{(k)} = 0$.

\vspace{-0.5em}
\subsubsection{Encoding Time Periodicity} 
Additionally, inspired by \cite{kazemi2019time2vec}, in order to capture the periodicity patterns in the time sequence within a work shift, we transform the scalar notion of time $t\subs$ into a vector representation, $\stampEmb\subs \in \mathbb{R}^d$
\begin{equation}
\label{eq:stamp}
    \stampEmb\subs [j] = \left\{
    \begin{array}{ll}
        \omega_j t\subs + \varphi_j, & \text{if  } j=1 \\
        \sin(\omega_j t\subs + \varphi_j), & \text{if  } 1<j\leq d 
    \end{array}
\right.
\end{equation}
where $\stampEmb\subs [j]$ is the $j$-th entry of the vector $\stampEmb\subs$, and $\omega_j$ and $\varphi_j$ are trainable frequency and bias variables. The first entry in $ \stampEmb$ captures the aperiodic pattern while the rest of the entries captures the periodic pattern by a sinusoidal function. 

\vspace{-0.5em}
\subsubsection{Action-time Joint Embedding}
We obtain the time-dependent activity embedding by concatenating the vectors of the action embedding, time interval embedding, and time periodicity embedding
\begin{equation}\label{tab:joint}
    \emb\subs = \text{Concat}([\actionEmb\subs; \intervEmb\subs; \stampEmb\subs])
\end{equation}
Alternatively, the aggregation method $\text{Concat}(\cdot)$ can be replaced by addition. We find concatenation works better for this dataset. The joint embedding $\emb\subs$ can then be used as the input to a sequence model for burnout prediction.
With the physician's activity logs encoded into the representation in Eq. (\ref{tab:joint}), in theory we can apply any single-level sequence model on top of the activity embedding layers for burnout prediction.


\subsection{Hierarchical Sequence Model}

To capture the temporal clustering patterns of physician activities and the natural shift-month hierarchical structure while improving computational efficiency, we propose \ours, a two-level model that naturally mirrors the hierarchical structure of sequential data.



\subsubsection{Low-level Encoder}
Given the sequence of action-time joint embedding of the $k$-th shift, $\mathbf{G}^{(k)} = [\emb\subs]_{i=1}^{N_k}$, we use an efficient sequence model (e.g., TCN, FCN) as the low-level encoder to obtain daily representations of physician behaviors, then we aggregate the output sequence by transforming it to a single vector
\begin{equation}\label{eq:low}
\begin{array}{cc}
     \textbf{h}^{(k)} = \Phi\Big(\mathbf{G}^{(k)}; \boldsymbol\theta_{E}\Big)
\end{array}
\end{equation}
where $\boldsymbol\theta_{E}$ denotes the parameters of low-level encoder, and $\shift^{(k)}$ denotes the daily representations. 

\subsubsection{High-level Encoder}
Similar to Eq. (\ref{eq:interv}) and Eq. (\ref{eq:stamp}) we first get high-level time interval embedding $\mathbf{p}^{(k)}$ and periodicity embedding $\mathbf{q}^{(k)}$ for the start time of the $k$-th shift, $t_1^{(k)}$. Then we condition the daily representation $\shift^{(k)}$ on the time embedding, and 
obtain the time-dependent representation of the $k$-th shift as $\mathbf{r}^{(k)} = \text{Concat}([\mathbf{h}^{(k)}; \mathbf{p}^{(k)}; \mathbf{q}^{(k)}])$.
Next, we use LSTM as the high-level encoder to aggregate the representation of each work shift into the work measurement for the whole survey month
\begin{equation}
\begin{array}{cc}
     {\mon}^{(k)} = \text{LSTM}\Big(\overrightarrow{\mon}^{(k-1)}, \mathbf{r}^{(k)}; \boldsymbol\theta_{H}\Big)
\end{array}
\end{equation}
where $\boldsymbol\theta_{H}$ denotes the parameters of high-level model LSTM, and $\mon^{(k)}$ denotes the cumulative activity representation. 
We use the last cumulative representation $\mon^{(T)}$ as the activity measurement for the whole survey month, where $T$ denotes the number of work shifts in the month.
Then we map the monthly representation $\mon^{(T)}$ to estimate the burnout label $y$ using multi-layer perceptron (MLP) as the classifier
\begin{equation}
    \gamma = \text{Softmax}\big(\text{MLP}\big(\mon^{(T)}; \boldsymbol\theta_{C}\big) \big)
\end{equation}
where $\gamma$ is the monthly risk score. The network parameters including the embedding parameters $\mathbf{W}_a$, $\mathbf{W}_b$, $\mathbf{d}_b$, $\boldsymbol{\omega}$ and $\boldsymbol{\varphi}$, the low-level model parameters $\boldsymbol\theta_{E}$, the high-level model parameters $\boldsymbol\theta_{H}$, and the classifier parameters $\boldsymbol\theta_{C}$,        can all be jointly trained by minimizing the cross entropy loss between true and estimated labels. The cross entropy (CE) loss can be written as 
\begin{equation}
    \mathcal{L}_H = \text{CE}(\gamma, y) = y\log \gamma + (1 - y)\log(1 - \gamma)
\end{equation}
Please refer to the Supplementary Sections for more details about the design choice of sequence encoder $\Phi$.

\subsection{Temporal Regularization}
\subsubsection{Temporal Consistency}
A potential challenge with the proposed hierarchical architecture is that the network must learn to pass input information across many low-level sequence steps and two-level sequence models in order to affect the output.
Inspired by RNN target replication \cite{lipton2015critical}, we propose Temporal Consistency (TC) regularization to apply an additional loss $\mathcal{L}_L$ for the low-level encoder of \ours. In TC, we use a linear scoring layer to measure the daily risk w.r.t. each shift:
\begin{equation}\label{eq:TCR}
\begin{split}
    \alpha^{(k)} &= \text{Softmax}(\mathbf{W}_R\cdot\mathbf{r}^{(k)} + \mathbf{d}_R) \\
    \mathcal{L}_L &= \frac{1}{T} \sum_{k=1}^{T} \text{CE}\big( \alpha^{(k)}, y\big)
\end{split}
\end{equation}
where $\mathbf{W}_R$ and $\mathbf{d}_R$ are trainable parameters, $S$ is the number of shifts, $\alpha^{(k)}\in[0, 1]$ denotes the daily risk of the $k$-th shift, and $\mathcal{L}_L$ denotes the TC loss that penalizes the overall mismatch between daily risks $[\alpha^{(1)}, ..., \alpha^{(T)}]$ and the monthly burnout outcome $y$.
Then the network can be trained by minimizing the composite loss 
\begin{equation}
\label{eq:loss}
    \mathcal{L} = \mathcal{L}_H + \lambda\mathcal{L}_L 
\end{equation}
where $\lambda$ is a trade-off parameter balancing the effect of accumulative monthly risk and overall daily risks. 
We select the value of $\lambda$ from $\{10^{-4}, 10^{-3}, ..., 10^{4}\}$, and find that $\lambda = 10^{-1}$ works the best for our the activity log dataset.

The TC regularization has three major effects. First, it helps pass the supervision signal directly to the low-level encoder for learning better class-specific representations. 
Distinct from target replication \cite{lipton2015learning} that replicates labels to the LSTM output at all steps, TC bypasses the LSTM-based high-level encoder and directly regularizes the low-level representation learning with higher flexibility.
Second, it regularizes the network to maintain relative temporal consistency across shifts by minimizing the overall mismatch between daily risks and the burnout outcome.
The intuition is that for most cases the clinical behaviors (workload and workflow) of a physician within a monthly rotation may remain similar across different shift days.
Third, it enables better model interpretability of \ours.
During inference time, while the monthly risk $\gamma$ is used to estimate risk of burnout, the daily risks $[\alpha^{(1)}, \alpha^{(2)}, ..., \alpha^{(T)}]$ in Eq. (\ref{eq:TCR}) can be used to reflect the dynamics of daily measurement in that month.

\subsubsection{Stochastic Tail Drop}
During model training, the activity logs of the whole month until the time of survey submission are mapped to the binary burnout labels.
A potential problem for model training is that, the time between end of the clinical rotation and survey submission may vary by hours to a few days for different individuals in different months. Thus the latest activity logs may describe the work of a new monthly rotation with potential significantly different workflow and workload, which can confuse the predictive model.
An ideal predictive model should not be overly dependent on the length of data and when to run model inference.
To seek more robust prediction, we propose {Stochastic Tail Drop} that stochastically drops out the sequence tail with a length randomly drawn from a fixed distribution at each iteration during training. Formally, we draw the length of dropped tails from a power distribution
\begin{equation}
    L\sim p(L) \propto (L_{max} - L)^\rho, \ \ 0 \leq L \leq L_{max}
\end{equation}
where $L_{max}$ is the maximum number of days allowed to drop and $\rho$ controls the distortion of distribution. We set $L_{max} = 5$ and $\rho = 2$ for all \ours variants.

\subsection{Semi-supervised \ours}
A challenge in burnout studies is the difficulty to collect surveys from physicians. As a result, only about half of activity logs are associated with burnout labels in our study. In contrast, activity logs are regularly collected for all physicians. To exploit the unlabeled activity logs, we design a semi-supervised framework that learns from all recorded activity logs and allows the generalizable knowledge transfer from the unlabeled data to the supervised predictive model \cite{liu2017semi}.
We adopt an unsupervised sequence autoencoder (Seq-AE) that learns to reconstruct the action-time sequence of each work shift, where the encoder shares the same model parameters with the low-level encoder $\Phi$ in Eq. (\ref{eq:low}). We adopted an appropriate sequence decoder $\Psi$ that mirrors the encoder $\Phi$ accordingly (e.g., use TCN as decoder for a TCN encoder),
\begin{equation}\label{eq:decoder}
    \textbf{S}^{(k)} = \Psi\Big(\textbf{H}^{(k)};\boldsymbol{\theta}_D\Big)
\end{equation}
where $\textbf{H}^{(k)} = [\textbf{h}^{(k)}, \textbf{h}^{(k)}, ..., \textbf{h}^{(k)}]$ replicates the daily representation $\textbf{h}$ to every activity step in the $k$-th work shift, and $\textbf{S}^{(k)} = [\textbf{s}^{(k)}_1, \textbf{s}^{(k)}_2, ..., \textbf{s}^{(k)}_{N_k}]$.
Then we reconstruct the sequence of actions $\textbf{e}$ and time stamps $\textbf{t}$: 
\begin{equation}
\begin{split}
    \hat{e}_i &= \text{Softmax}(\text{ReLU}(\textbf{W}_e \textbf{s}_i^{(k)} + \textbf{d}_e)) \\
    \hat{t}_i &= \text{Softmax}(\text{ReLU}(\textbf{W}_t \textbf{s}_i^{(k)} + \textbf{d}_t))
\end{split}
\end{equation}
where $\textbf{W}$ and $\textbf{d}$ are weight and bias parameters. The action projection matrix $\textbf{W}_e$ is initialized with the transpose of the pre-trained embedding matrix $\textbf{W}_a$ for quicker convergence.
Then the encoder parameters $\boldsymbol{\theta}_L$ and decoder parameters $\boldsymbol{\theta}_R$, $\textbf{W}$ and $\textbf{d}$ can be pre-trained by minimizing the cross entropy 
\begin{equation}\label{eq:loss_u}
    \mathcal{L}_u = \text{CE}(\hat{\textbf{e}}, \textbf{e}) +  \text{CE}(\hat{\textbf{t}}, \textbf{t})
\end{equation}
The Seq-AE is pre-trained on all available activity logs (labeled or unlabeled) and then transferred to and reused by \ours as the low-level encoder and fine-tuned with the predictive model on labeled data
(see Appendix for design details of Seq-AE).
 

%% file: HierTCN-KDD22/experiment.tex

\section{Experiment}
\label{sec:experiment}

\subsection{Setup}

The primary focus of our analysis was to develop generalizable predictive models that could be translated for predicting burnout outcomes for new unseen participants. Towards this end, the training and testing data are split based on participants, where no activity logs (from different months) of any participant simultaneously exist in both training and testing set. Considering the relatively small sample size (i.e., number of valid surveys), we evaluate each method with \textit{repeated 5-fold cross-validation} (CV) in order to get as close estimation as possible to the true out-of-sample performance of each model on unseen individuals. For each fold, the whole dataset is split into 80\% training set (10\% training data used for validation) and 20\% testing set. We repeat the cross-validation with different random split for 6 rounds and report the mean and standard deviation of CV results.
We use {accuracy}, {area under the receiver operating characteristic} (AUROC), and {area under the precision-recall curve} (AUPRC) as the metric measures to evaluate the burnout prediction performance.
All non-neural models were implemented using Scikit-learn
1.0.1 with Python, and all deep learning models were implemented using TensorFlow
2.6.0. All models were tested on Linux Ubuntu 20.04 empowered by Nvidia RTX 3090 GPUs.


\subsection{Baseline Methods}
We compare our proposed burnout prediction framework to the following baseline methods. 
\vspace{-0.3em}
\begin{itemize}
    \item \textbf{GBM}/\textbf{SVM}/\textbf{RF}: Gradient Boosting Machines implemented with XGBoost \cite{chen2016xgboost}, Support Vector Machines, and Random Forests, used in \cite{lou2021predicting} for burnout prediction. We follow \cite{lou2021predicting} to extract a set of summary statistics of activity logs as features for prediction.
    
    \item \textbf{FCN} \cite{wang2017time}: Full Convolutional Networks, a deep CNN architecture with Batch Normalization, shown to have outperformed multiple strong baselines on 44 benchmarks for time series classification.
    
    \item \textbf{\simtcn} \cite{lea2017temporal}: A primitive architecture of TCN with fixed dilation and Max Pooling layers used for videos \cite{lea2017temporal}. 
    
    \item \textbf{\restcn} \cite{bai2018empirical}. A popular TCN architecture with exponentially enlarged dilation. The cross-layer residual connections enables the construction of a much deeper network \cite{bai2018empirical}.
    
    \item \textbf{H-RNN} \cite{zhao2017hierarchical}: Hierarchical RNN, a multi-level LSTM model used for long video classification \cite{zhao2017hierarchical}.
    
    \item \textbf{\higru}: Hierarchical baselines, implemented with GRU as the \ours low-level encoder.
    
    \item \textbf{Semi-\restcn}: A semi-supervised single-level model baseline, implemented with \restcn. We pre-train the single-level TCN with an TCN-AE.
\end{itemize}
{All the compared deep models were implemented with our proposed time-dependent activity embedding for activity logs.}
As variants of our proposed hierarchical framework, \textbf{\hifcn}, \textbf{\hisimtcn}, and \textbf{\hirestcn} corresponds to \our-based predictive model with the low-level encoder $\Phi$ instantiated by FCN, \simtcn, and \restcn. Similar for the semi-supervised model variants.

\setlength{\tabcolsep}{3.5pt}
\begin{table}[t]
  \caption{Random repeated cross-validation results.}
\centering
\fontsize{8.5pt}{10pt}\selectfont
\begin{tabular}{||lccc||}

\hline

  {\color[HTML]{000000} \textsc{Method}} &
  {\color[HTML]{000000} AUROC} &
  {\color[HTML]{000000} AUPRC} &
  {\color[HTML]{000000} \textsc{Accuracy}} \\ 

\hline
\hline

  {\color[HTML]{000000} GBM} &
  {\color[HTML]{000000} .5597 (.0214)} &
  {\color[HTML]{000000} .4646 (.0676)} &
  {\color[HTML]{000000} .5717 (.0361)}  \\

  {\color[HTML]{000000} SVM} &
  {\color[HTML]{000000} .5793 (.0290)} &
  {\color[HTML]{000000} .4683 (.0513)} &
  {\color[HTML]{000000} .5545 (.0326)} \\

  {\color[HTML]{000000} RF} &
  {\color[HTML]{000000} .5645 (.0243)} &
  {\color[HTML]{000000} .4647 (.0606)} &
  {\color[HTML]{000000} .5611 (.0355)} \\ 
  
\hline
\hline

  {\color[HTML]{000000} FCN \cite{wang2017time}} &
  {\color[HTML]{000000} .6001 (.0141)} &
  {\color[HTML]{000000} .5167 (.0167)} &
  {\color[HTML]{000000} .6099 (.0157)} \\
  
  {\color[HTML]{000000} \simtcn \cite{lea2017temporal}} &
  {\color[HTML]{000000} .5724 (.0313)} &
  {\color[HTML]{000000} .4794 (.0373)} &
  {\color[HTML]{000000} .5835 (.0288)} \\

  {\color[HTML]{000000} \restcn \cite{bai2018empirical}} &
  {\color[HTML]{000000} .6171 (.0258)} &
  {\color[HTML]{000000} .5284 (.0602)} &
  {\color[HTML]{000000} .6150 (.0232)} \\

\hline
\hline
  
  {\color[HTML]{000000} H-RNN \cite{zhao2017hierarchical}} &
  {\color[HTML]{000000} .5935 (.3780)} &
  {\color[HTML]{000000} .4774 (.0621)} &
  {\color[HTML]{000000} .6012 (.0323)} \\

  {\color[HTML]{000000} \higru} &
  {\color[HTML]{000000} .5871 (.0155)} &
  {\color[HTML]{000000} .4740 (.0457)} &
  {\color[HTML]{000000} .6117 (.0277)} \\
  
  {\color[HTML]{000000} \hifcn} &
  \textbf{\color[HTML]{000000} .6358 (.0220)} &
  \underline{\color[HTML]{000000} .5588 (.0336)} &
  \textbf{\color[HTML]{000000} .6449 (.0200)} \\
  
  {\color[HTML]{000000} \hisimtcn} &
  \underline{\color[HTML]{000000} .6347 (.0181)} &
  {\color[HTML]{000000} .5502 (.0191)} &
  \underline{\color[HTML]{000000} .6400 (.0218)} \\
  
  {\color[HTML]{000000} \hirestcn} &
  {\color[HTML]{000000} .6244 (.0295)} &
  \textbf{\color[HTML]{000000} .5611 (.0691)} &
  {\color[HTML]{000000} .6390 (.0064)} \\

\hline
\hline

  {\color[HTML]{000000} Semi-ResTCN} &
  {\color[HTML]{000000} .6185 (.0199)} &
  {\color[HTML]{000000} .5454 (.0378)} &
  {\color[HTML]{000000} .6170 (.0205)} \\

  {\color[HTML]{000000} Semi-\hifcn} &
  \textbf{\color[HTML]{000000} .6479 (.0189)} &
  \textbf{\color[HTML]{000000} .5680 (.0236)}&
  {\color[HTML]{000000} .6383 (.0191)}\\

  {\color[HTML]{000000} Semi-\hisimtcn} &
  {\color[HTML]{000000} .6400 (.0204)} &
  \underline{\color[HTML]{000000} .5559 (.0151)}&
  \underline{{\color[HTML]{000000} } .6428 (.0229)}\\

  {\color[HTML]{000000} Semi-\hirestcn} &
  \underline{\color[HTML]{000000}  .6312 (.0299) }&
  \underline{}{\color[HTML]{000000} .5536 (.0479)} &
  \textbf{\color[HTML]{000000} .6450 (.0270)} \\
  

  
\hline
\end{tabular}
\label{tab:prediction}
\end{table}

\subsection{Experiment Results}
\subsubsection{Overall Performance}
Table \ref{tab:prediction} summarizes the performance of all the models, including non-deep-learning models, single-level sequence models, hierarchical models, and semi-supervised models.
Our proposed \ours framework including its semi-supervised extension outperforms all the baseline methods, achieving \textbf{5.0\%/7.5\%} improved average AUROC/AUPRC scores over the best deep learning baseline model (\restcn \cite{bai2018empirical}) and \textbf{11.9\%/21.3\%} improved average AUROC/AUPRC over the state-of-the-art activity log based burnout prediction approach \cite{lou2021predicting}.

\vspace{-0.5em}
\subsubsection{Feature Engineering vs. Representation Learning}
In general, the non-deep models GBM, SVM, and RF 
show inferior prediction performance compared to 
deep learning models. This may result from the less effective capacity of simple statistical features (e.g., EHR time, number of notes reviewed) in capturing complex activity dynamics and temporality.
In contrast, armed with our proposed activity embedding, the deep models can learn deep representations capable of encoding more complicated patterns.

\vspace{-0.5em}
\subsubsection{Single-level vs. Hierarchical}
Among the single-level sequence models, ResTCN achieves the best performance due to its effective deep architecture with residual mechanism and delicately designed convolutional blocks.
The corresponding \ours extension improves the performance of each base model, respectively. Especially for \simtcn, performing the worst among the single-level baseline models due to its relatively primitive architecture, \hisimtcn achieves 10.9\%/14.8\% average improvement on AUROC/AUPRC. 
The steady improvement regardless of the base model shows that the hierarchical architecture tailored for the problem enables \ours to better capture the multi-level structure in activity logs and complex temporality and dynamics.

\vspace{-0.5em}
\subsubsection{Supervised vs. Semi-supervised}
Compared to the supervised \ours models, in general all the three Semi-\ours counterparts achieve better average performance (except that Semi-\hirestcn has slightly worse AUPRC than \hirestcn). We can see that based on the Seq-AE pre-training, our semi-supervised framework is able to effectively extract generalizable patterns from unlabeled activity logs and transfer knowledge to \ours. This sheds light to potential improved prediction efficacy in real-world clinical practice
when the costly burnout labels are limited in number but the huge amount of unlabeled activity logs are available.

\setlength{\tabcolsep}{3.5pt}
\begin{table}[t]
  \caption{Model size and average training time per epoch ($\star\star$ indicates recurrent models with no parallel acceleration).}
\centering
\fontsize{8.5pt}{11pt}\selectfont
\begin{tabular}{||l|c|cc||}

\hline
  {\color[HTML]{000000} \textsc{Model}} &
  {\color[HTML]{000000} \# \textsc{Params}} &
  {\color[HTML]{000000} \textsc{Pre-train}} &
  {\color[HTML]{000000} \textsc{Training}} \\ 
  
\hline\hline
  
    {\color[HTML]{000000} FCN \cite{wang2017time}} &
  {\color[HTML]{000000} 1,060 K} &
  {\color[HTML]{000000} -} &
  {\color[HTML]{000000} 12.6 s} \\
  
  {\color[HTML]{000000} \simtcn \cite{lea2017temporal}} &
  {\color[HTML]{000000} 1,428 K} &
    {\color[HTML]{000000} -} &
  {\color[HTML]{000000} 7.8 s} \\

  {\color[HTML]{000000} \restcn \cite{bai2018empirical}} &
  {\color[HTML]{000000} 1,887 K} &
    {\color[HTML]{000000} -} &
  {\color[HTML]{000000} 59.8 s} \\ 
  
\hline\hline

  {\color[HTML]{000000} H-RNN \cite{zhao2017hierarchical}} &
  {\color[HTML]{000000} 694 K} &
    {\color[HTML]{000000} -} &
  {\color[HTML]{000000} 314.7 s} \\ 
  
  {\color[HTML]{000000} \higru**} &
  {\color[HTML]{000000} 635 K} &
    {\color[HTML]{000000} -} &
  {\color[HTML]{000000} 304.2 s} \\ 
  

  {\color[HTML]{000000} \hifcn} &
  {\color[HTML]{000000} 814 K} &
    {\color[HTML]{000000} -} &
  {\color[HTML]{000000} 12.4 s} \\

  {\color[HTML]{000000} \hisimtcn} &
  {\color[HTML]{000000} 880 K} &
    {\color[HTML]{000000} -} &
  \underline{\color[HTML]{000000} 6.3 s} \\
  
  {\color[HTML]{000000} \hirestcn} &
  {\color[HTML]{000000} 1,233 K} &
    {\color[HTML]{000000} -} &
  {\color[HTML]{000000} 49.6 s} \\

\hline\hline

    {\color[HTML]{000000} Semi-ResTCN}  &
  {\color[HTML]{000000} 2,574 K} &
    {\color[HTML]{000000} 357.0 s} &
  {\color[HTML]{000000} 51.2 s} \\
  
{\color[HTML]{000000} Semi-\hifcn} &
  {\color[HTML]{000000} 1,455 K} &
    {\color[HTML]{000000} 81.0 s} &
  {\color[HTML]{000000} 11.4 s} \\
  
  {\color[HTML]{000000} Semi-\hisimtcn} &
  {\color[HTML]{000000} 1,216 K} &
    \textbf{\color[HTML]{000000} 65.7 s} &
  \textbf{\color[HTML]{000000} 5.9 s} \\
  
  {\color[HTML]{000000} Semi-\hirestcn} &
  {\color[HTML]{000000} 2,209 K} &
    {\color[HTML]{000000} 344.7 s} &
  {\color[HTML]{000000} 33.2 s} \\

\hline\hline

    {\color[HTML]{000000} GRU**} &
  {\color[HTML]{000000} 466 K} &
    {\color[HTML]{000000} -} &
  {\color[HTML]{000000} > 5 h} \\
  
    {\color[HTML]{000000} LSTM**} &
  {\color[HTML]{000000} 550 K} &
    {\color[HTML]{000000} -} &
  {\color[HTML]{000000} > 5 h} \\

\hline
\end{tabular}
\label{tab:time}
\end{table}

\vspace{-0.5em}
\subsubsection{Computational Efficiency}
Table \ref{tab:time} summarizes the model complexity and training time. 
Our proposed \ours framework is able to train with high efficiency, spending just a few seconds for one epoch on over 6 million activity logs (training set). 
In contrast, it takes hours to finish one epoch for LSTM and GRU on our dataset, which makes them unsuitable for this problem. Hence we do not report the performance in Table \ref{tab:prediction} due to extremely high time cost.
Despite the hierarchical structure, RNN-based models H-RNN and HierGRU still run about $50\times$ slower than \hisimtcn.
All the three \ours variants maintain comparable training speed of their base models. 


\setlength{\tabcolsep}{3.5pt}
\begin{table}[t]
  \caption{Effect of each regularization mechanism and embedding choice on \hisimtcn.}
\centering
\fontsize{8.5pt}{11pt}\selectfont
\begin{tabular}{||lccc||}
\hline
  {\color[HTML]{000000} \textsc{Variant}} &
  {\color[HTML]{000000} AUROC} &
  {\color[HTML]{000000} AUPRC} &
  {\color[HTML]{000000} \textsc{Accuracy}} \\ 
\hline
\hline

  {\color[HTML]{000000} w/o TC} &
  {\color[HTML]{000000} .6041 (.0100)} &
  {\color[HTML]{000000} .5493 (.0394)} &
  {\color[HTML]{000000} .5952 (.0156)} \\

  {\color[HTML]{000000} w/o Tail Drop} &
  {\color[HTML]{000000} .6251 (.0135)} &
  {\color[HTML]{000000} .5343 (.0147)} &
  {\color[HTML]{000000} .6236 (.0074)} \\ 
  
  
\hline\hline

  {\color[HTML]{000000} w/o Pretrain} &
  {\color[HTML]{000000} .6098 (.0257)} &
  {\color[HTML]{000000} .5273 (.0365)} &
  {\color[HTML]{000000} .6276 (.0129)} \\ 

  {\color[HTML]{000000} w/o Interval} &
  {\color[HTML]{000000} .6171 (.0122)} &
  {\color[HTML]{000000} .5340 (.0194)} &
  {\color[HTML]{000000} .6311 (.0420)} \\

  {\color[HTML]{000000} w/o Periodicity} &
  {\color[HTML]{000000} .6166 (.0300)} &
  {\color[HTML]{000000} .5216 (.0222)} &
  {\color[HTML]{000000} .6370 (.0141)} \\ 
  
  {\color[HTML]{000000} Concat $\rightarrow$ Add} &
  {\color[HTML]{000000} .6237 (.0240)} &
  {\color[HTML]{000000} .5393 (.0262)} &
  {\color[HTML]{000000} .6328 (.0232)} \\ 
  
\hline\hline
 {\color[HTML]{000000} w/ All} &
  \textbf{\color[HTML]{000000} .6347 (.0181)} &
  \textbf{\color[HTML]{000000} .5502 (.0191)} &
  \textbf{\color[HTML]{000000} .6400 (.0218)} \\
  
\hline
\end{tabular}\label{tab:ablation}
\end{table}

\subsection{Performance Analysis}

\begin{figure}[t]
    \centering
    \includegraphics[width=1.\linewidth]{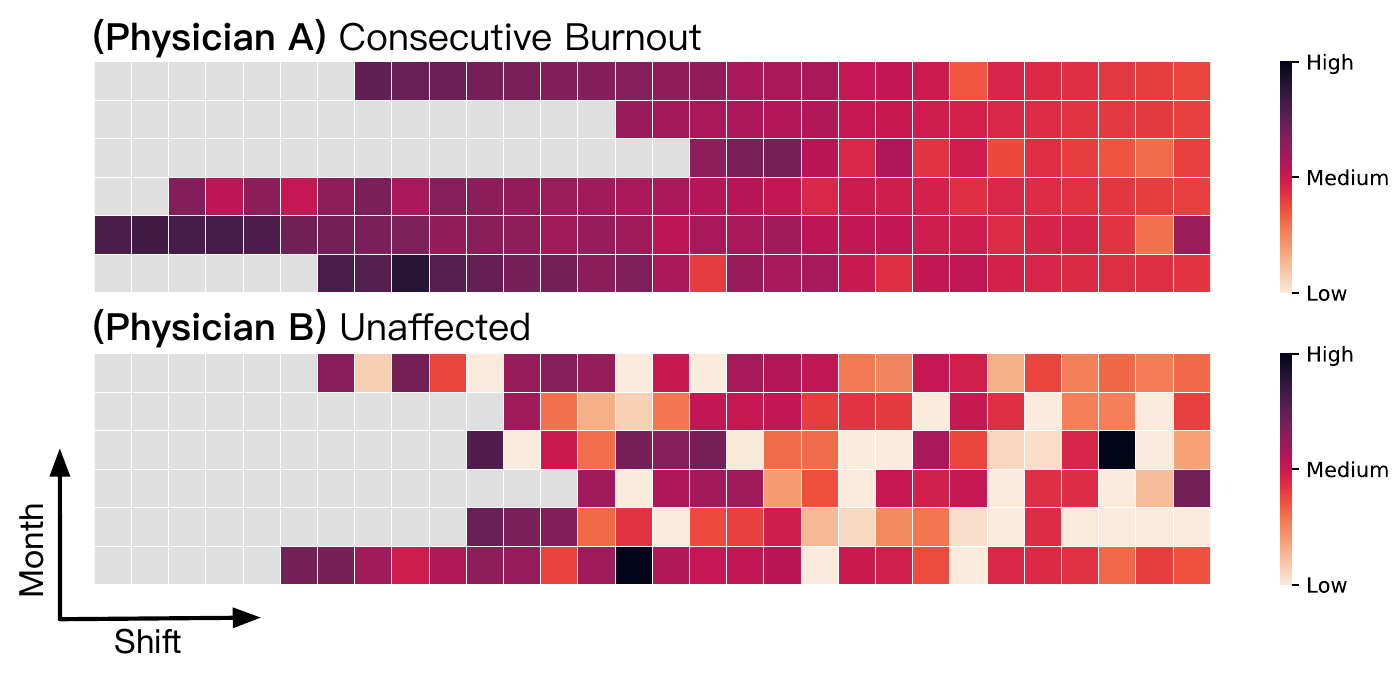}
    \caption{Visualized daily risks $\alpha$ of two typical physician participants over all shifts (horizontal axis, aligned to the right) across all 6 months (vertical axis). See Appendix for more examples.}
    \label{fig:softmax}
\end{figure}

\begin{figure}[t]
    \centering
    \vspace{-1em}
    \includegraphics[width=0.68\linewidth]{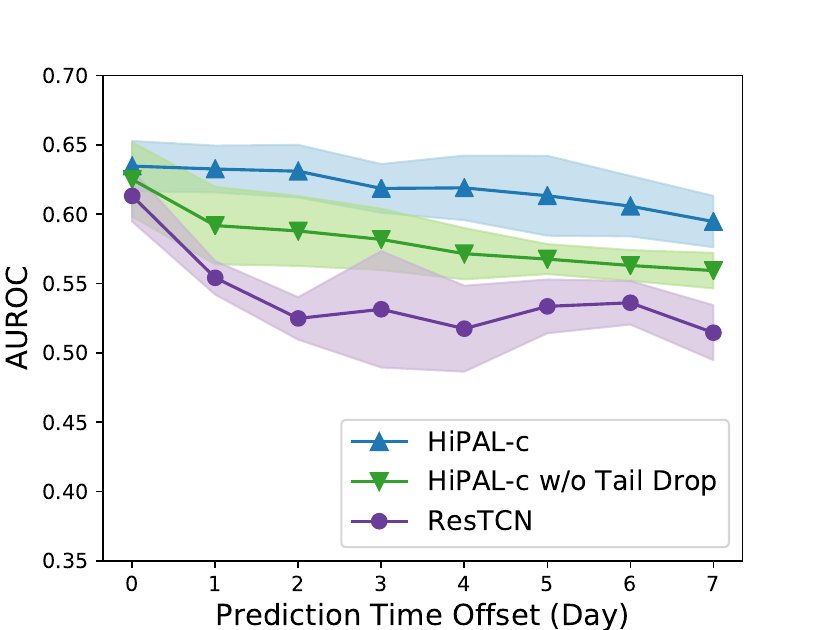}
    \caption{Effect of Stochastic Tail Drop on model robustness against uncertainty of prediction time.}
    \label{fig:offset}
\end{figure}

\subsubsection{Effects of Temporal Regularization}
As shown in Figure \ref{tab:ablation}, both the TC and Tail Drop regularization help improve the prediction performance.
We further test the effect of Stochastic Tail Drop on model robustness to uncertainty of time and data lengths. Figure \ref{fig:offset} shows the performance variance with various prediction time offset (by days) that corresponds to different lengths of input activity logs. 
Compared to the \ours variants, the AUROC of the best single-level model \restcn declines more drasticly. The Tail Drop enables \ours to have slower AUROC decline over larger time offset.
Without Tail Drop, the AUROC of \hisimtcn quickly drops to below 0.6 with only 1-day offset while the full model still maintains an AUROC of 0.6 with 7-day offset. We can see that Tail Drop can improve the robustness against uncertainty of model inference time, making the model less dependent to the data lengths.

\subsubsection{Choice of Embedding}
Table \ref{tab:ablation} summarizes the ablation study based on \hisimtcn.
Among different embedding configurations, we can see that our current \ours design that incorporates the pre-trained action embedding, time interval embedding and time periodicity embedding performs the best.

\subsubsection{Visualization of Daily Risks}
Figure \ref{fig:softmax} shows the visualized daily risks $\alpha$ (Eq. (\ref{eq:TCR})) across all 6 months of two typical physician participants from the testing set, one being consecutively burned-out in every month and the other staying unaffected. The daily risk scores can reflect the dynamics of daily measurement across shifts of a month. 
We can observe that the physicians with the two typical types of wellness status show visually distinct daily risk patterns (see Appendix for more examples).
The daily risks of Physician A continuously remain at a medium to high level across shifts in each month.
In contrast, Physician B seems to have various levels of daily risk that changed between low and high intermittently. This may contribute to the reduction of the cumulative monthly risk of burnout.
We can see that the TC regularizer with daily risk measures allows \ours to provide interpretable burnout prediction. 

\begin{table}[t]
    \centering
    \fontsize{8pt}{10pt}\selectfont
    \caption{Trade-off between sensitivity and specificity.}
    \begin{tabular}{||c c c c c||}
    \hline
\multirow{2}{*}{\textsc{Scenario}} & \multicolumn{2}{c}{\textsc{Cost}} & \multirow{2}{*}{\textsc{Sensitivity}} & \multirow{2}{*}{\textsc{Specificity}} \\ \cline{2-3}
 & \multicolumn{1}{c}{\textsc{Burnout}} & \textsc{Intervention} &  &  \\ 
 \hline\hline
 A & \multicolumn{1}{c}{Higher} &  Lower  & .8000 & .4054 \\
 B & \multicolumn{1}{c}{Lower}  & Higher  & .4782 & .8000 \\ 
    \hline
    \end{tabular}
    \label{tab:case}
\end{table}

\subsection{Effectiveness and Potential Impact}
High levels of burnout can lead to medical errors, substance abuse, and suicidal ideation \cite{hu2019discrimination}, so preventing or mitigating burnout has meaningful public health benefits.
Currently, due to the difficulty of assessing physician wellness in real time, the interventions for physician burnout mainly focus on organizational and system-level interventions, such as improving workflow design and enhancing teamwork and communication \cite{wiederhold2018intervention}. 
An end-to-end burnout monitoring and prediction system like \ours can offer new potential for real-time burnout phenotyping and personalized individual-level interventions.

Different individual-level interventions for physician burnout might have different costs, financially and administratively. For different situations, we may require a different sensitivity (true positive rate) and specificity ($1 - \text{false positive rate}$) for the burnout predictive model, and thus the model output must be tuned accordingly.
As the best variant, Semi-\hifcn achieves an AUROC of 0.6479, which reflects the average sensitivity over all specificity levels. 
Table \ref{tab:case} shows the model performance under two different practical situations. 
Scenario A represents the situation with a low-cost intervention, such as EHR training and online cognitive therapy, where usually a predictive model with high sensitivity -- detecting most cases of true burnout -- is preferred. With the sensitivity set as 0.8, our model presents a moderate specificity of 0.4054, meaning that nearly 60\% of unaffected physicians would be included in the personal interventions. This is acceptable since these interventions also benefit the physicians with normal wellness status and can help prevent future burnout.
Scenario B represents the situation with a high-cost intervention, such as scheduling additional days off or vacations. This is when a more specific predictive model is preferred. With the specificity set as 0.8, our model still achieves a sensitivity of 0.4782, meaning that the prediction can still benefit nearly half of the burned-out physicians in this extreme.


This work is not without limitations. 
In clinical practice, not all physician work is EHR-related or tracked by the EHR system, and as such the physician behaviors reflected in the activity logs do not always align with the actual wellness outcome. This may bound the predictive performance of a model trained exclusively based on activity logs.
Constrained by the current extent of our study at the moment, only a limited number of burnout labels have been collected, which could have restrained the power of deep learning from being fully exploited. For future work, the availability of much greater amount of EHR activity log data may allow us to explore more advanced semi-supervised or transfer learning approaches for better burnout prediction to facilitate physician well being.


%% file: HierTCN-KDD22/conclusion.tex
\section{Conclusion}
In this paper, we presented \ours, the first end-to-end deep learning framework for predicting physician burnout based on physician activity logs available in any electronic health record (EHR) system. The \ours framework includes a time-dependent activity embedding mechanism tailored for the EHR-based activity logs to encode raw data from scratch. We proposed a hierarchical sequence learning framework to learn deep representations of workload that contains multi-level temporality from the large-scale clinical activities with high efficiency, and provide interpretable burnout prediction. The semi-supervised extension enables \ours to utilize the large amount of unlabeled data and transfer generalizable knowledge to the predictive model.
The experiment on over 15 million real-world physician activity logs collected from a large academic medical center shows the advantages of our proposed framework in predictive performance of physician burnout 
and training efficiency over state of the art approaches.

\section*{Acknowledgements}
This study was funded by the Fullgraf Foundation and the Washington University/BJC HealthCare Big Ideas Healthcare Innovation Award. Sunny S. Lou was supported by NIH 5T32GM108539-07.

%% file: HierTCN-KDD22/supplementary.tex
\begin{appendices}
\appendixpage

\balance

\section{Statistics of Dataset} \label{sec:more_details_of_data}

\begin{table}[h]
    \centering
    \fontsize{8pt}{10pt}\selectfont
    \caption{Overall statistics of dataset.}
    \begin{tabular}{lr}
    \toprule
    Statistics & Number\\
    \midrule
    \# total data points (actions) & 15,767,634 \\
    \# types of actions & 1,961 \\
    \# participants & 88 \\
    \# total months of activity logs & 754 \\
    \# months with eligible surveys (labels) & 391 \\
    \specialrule{0.01em}{1.5pt}{1.5pt} 
    \# total work shifts & 11,890 \\
    \# avg shifts per month per participant & 16 \\
    \bottomrule
    \end{tabular}
    \label{tab:dataset}
\end{table}

\begin{table}[h]
    \centering
    \fontsize{8pt}{10pt}\selectfont
    \caption{Statistics of data sequence grouped by participants ($p$), survey months ($m$), and work shifts ($s$).}
    \begin{tabular}{c|r|r|r}
    \toprule
    Group & by $ p$ & by $ p\cdot m$ & by $ p\cdot m\cdot s$\\
    \midrule
    \# sequences & 88 & 754 & 11,890\\ 
    avg length&  179,177 & 20,911 & 1,326\\
    std & 62,668 & 14,683 & 891 \\
    max length & 464,711 & 90,125 & 8,459\\
    \bottomrule
    \end{tabular}
    \label{tab:sequence}
\end{table}

\section{Design Choice of Sequence Model}
\label{sec:design}
In this section, we introduce the three single-level convolution-based models we adopted, \simtcn, \restcn, and FCN as the low-level sequence encoder (Eq. (\ref{eq:low})) in \hisimtcn, \hirestcn, and \hifcn, respectively. All these sequence model must work with our proposed activity embedding for burnout prediction based on activity logs.

\subsection{Fully Convolutional Networks}
Full Convolutional Networks (FCN), a deep CNN architecturewith Batch Normalization, has shown compelling quality and efficiency for tasks on images such as semantic segmentation. Later work \cite{wang2017time} applied TCN on time series classification, which has shown to have outperformed multiple strong baselines on 44 time series benchmarks. It also outperformed another widely used CNN-based model with residual connections -- ResNet \cite{wang2017time} -- on most of the above datasets. Hence, in this paper, we select FCN implemented by \cite{wang2017time} as the representative of conventional CNN-based sequence model for comparison and also as the one of the base model choices for our \ours framework.

An FCN model consists of several basic convolutional blocks. A basic block is a convolutional layer followed by a Batch Normalization layer and a ReLU activation layer, as follows:
\begin{equation}
\begin{split}
    \mathbf{y} & = \mathbf{W} * \textbf{x} + \textbf{d} \\
    \textbf{z} & = \text{BatchNorm}(\textbf{y}) \\
    \textbf{x}' & = \text{ReLU}(\textbf{z})
\end{split}
\end{equation}
where $*$ is the convolution operator. 
For \hifcn, we use 3 blocks for the FCN and use the filter sizes $\{128, 256, 128\}$ for each block. The kernels are set with sizes $\{8, 5, 3\}$.

\subsection{Temporal Convolutional Networks}

\begin{figure}[h]
    \centering
    \includegraphics[width=0.85\linewidth]{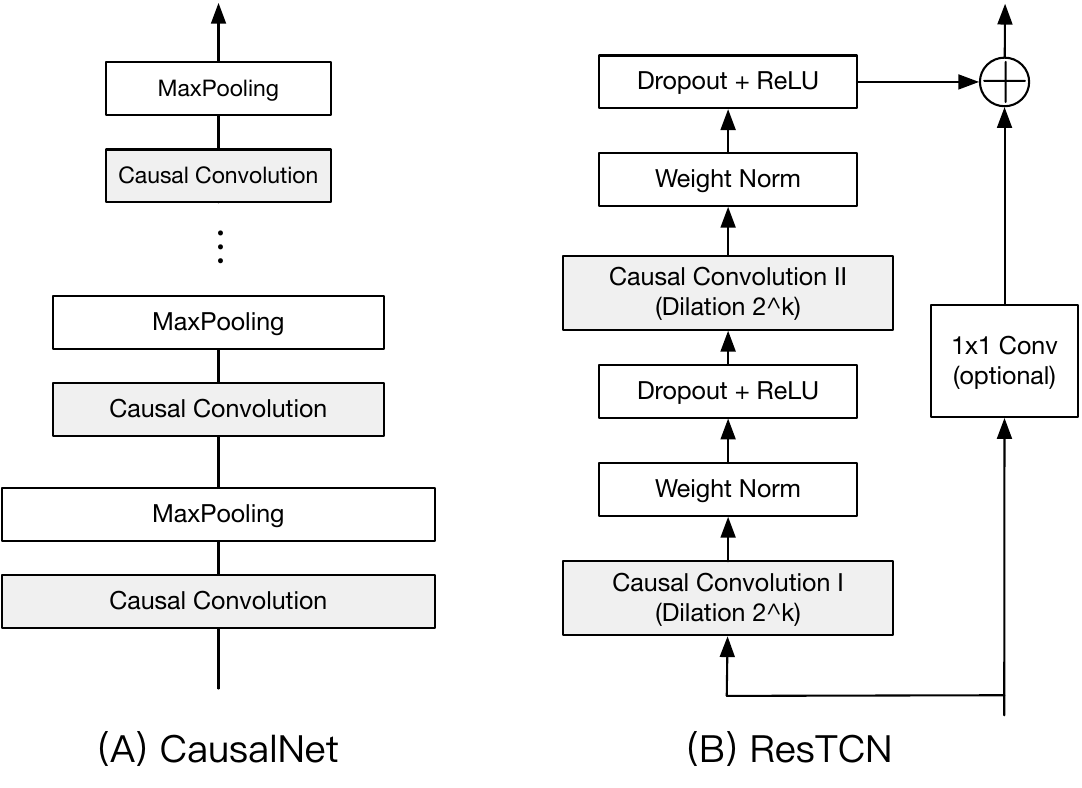}
    \caption{Architecture of \simtcn and \restcn.}
    \label{fig:tcn}
\end{figure}

Temporal convolutional networks (TCN) is a family of efficient 1-D convolutional sequence models where convolutions are computed across time \cite{bai2018empirical,lea2017temporal}. Different from the RNN family of sequence models, in TCN computations are performed layer-wise where every time-step is updated concurrently instead of recurrently \cite{lea2017temporal}.
TCN differs from dypical 1-D CNN mainly by using a different convolution mechanism, dilated causal convolution. 
Formally, for a 1-D sequence input $\mathbf{X}=[\textbf{x}_1, ..., \textbf{x}_T]\in\mathbb{R}^{d\times T}$ and a convolution filter $\mathbf{f}\in\mathbb{R}^{k\times d}$, the dilated causal convolution operation $F$ on element $t$ of the sequence is defined as
\begin{equation}
    F(\mathbf{x}_t) = (\mathbf{X} *_d \mathbf{f})(t) = \sum_{i=0}^{k-1} \textbf{f}_i^T\cdot \mathbf{x}_{t-d\cdot i}, \ \ s.t., \ t\geq k, \textbf{x}_{\leq 0}:= 0
\end{equation}
where $d$ is the dilation factor, $k$ is the filter size, and $t - d\cdot i$ accounts the past. Dilated convolution, i.e., using a larger dilation factor $d$, enables an output at the top level to represent a wider range of inputs, effectively expanding the receptive field of convolution. 
Causal convolution, i.e., at each step the convolution is only operated with previous steps, ensures that no future information is leaked to the past \cite{bai2018empirical}. This feature enables TCN to have similar directional structure as RNN models. 
Then the output sequence $\textbf{X}'\in\mathbb{R}^{k\times T}$ of the dilation convolution layer can be written as
\begin{equation}
    \textbf{X}' = [F(\mathbf{x}_1), F(\mathbf{x}_2), ..., F(\mathbf{x}_T)]
\end{equation}
Usually Layer Normalization or Batch Normalization regularization is applied after the convolutional layer for better performance \cite{lea2017temporal,bai2018empirical}. A TCN model is usually built with multiple causal convolutional layers with a wide receptive field that accounts for long sequence input.
There are two major variants of TCN, with their architecture shown in Figure \ref{fig:tcn}.

\begin{itemize}
    \item \textbf{\simtcn} \cite{lea2017temporal}:An early practice as in \cite{lea2017temporal} that connects multiple causal convolutional layers together using downsampling layers (e.g., Average Pooling) between each two convolutional layers as many CNN models do. With the help of the downsampling layers, a long sequence input can be progressively summarized into a lower-dimensional dense representation. 
    
    \item \textbf{\restcn} \cite{bai2018empirical}: A more popular and empirically effective TCN approach that applies residual connections (i.e., shortcuts) among dilated causal convolutional layers to further obtain deeper TCN. Instead of using downsampling layers, in \restcn the dilation $d$ is increased exponentially (e.g., using \{1, 2, 4, ...\} as the dilation factors) to realize exponentially large receptive field.
\end{itemize}

In \simtcn, the spatial scale (number of time steps) keeps reducing with higher layers by the Max Pooing layers, while in \restcn, the spatial scale keeps unchanged. 
For our prediction task, we have different configuration for \ours implemented with \simtcn and \restcn. 
For \simtcn, we use a Flattening layer as the final feature aggregation layer before the final Softmax layer. For \restcn, only the TCN output at the final step is used as the representation for prediction. In our implementation, both \simtcn and \restcn based \ours model have 6 causal convolutional layers. For single-level \simtcn and \restcn, we set the number of layers as 12 to increase the convolutional receptive field over much longer sequences.

\begin{figure*}[t]
    \centering
    \includegraphics[width=0.92\linewidth]{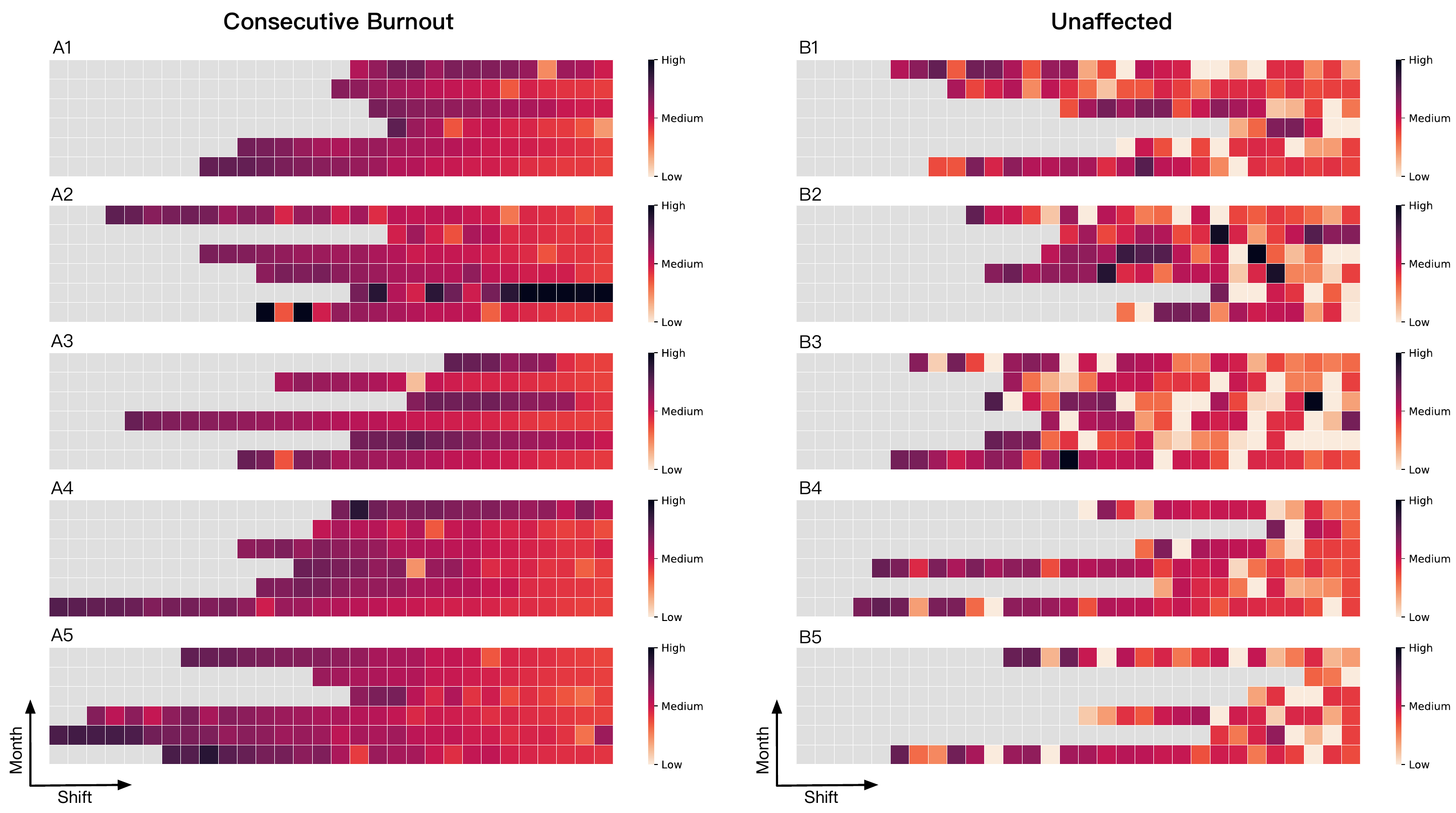}
    \caption{Visualized daily risks (Softmax output of low-level encoder) for two groups of typical physician participants over 6 months, a group that has been consecutively burned-out (left column) and the other stayed unaffected (right column). Darker colors correspond to larger Softmax score. Grey colors denote empty shifts (all shifts are aligned to the right).}
    \label{fig:softmax_full}
\end{figure*}

\section{Design of Sequence Autoencoder}
Based on the different architecture of the three base sequence models, FCN, \simtcn, and \restcn, we configure the Seq-AE in Eq. (\ref{eq:decoder}) differently for different base model.
For FCN, since there are no spacial size change at all, for the decoder in Eq. (\ref{eq:decoder}), we directly take the same FCN for the encoder in Eq. (\ref{eq:low}) but reverse the order of the layers to get a mirrored structure. For \simtcn as the encoder where downsampling layers (e.g., MaxPool) are used to reduce the spatial scale, in the decoder, we replace the downsampling layers to upsampling layers to increase spatial scale for data reconstruction. And for \restcn, since we usually use the last output of \restcn as the representation for any downstream task, in the decoder, we first replicated the representation produced by the encoder in Eq. (\ref{eq:low}) to every time steps and then feed them to the decoder counterpart of \restcn configured in the same way. 
Table \ref{tab:param} shows the hyperparameter used in the implementation.


\begin{table}[t]
    \centering
    \fontsize{8pt}{10pt}\selectfont
    \caption{Hyperparameters (hierarchical/single-level).}
    \begin{tabular}{||l l l l l l||}
    \hline
    Batch size & 2 & Learning rate & 0.001 & Optimizer & Adam\\
    Epochs     & 50 & Shifts       & 30/-  & Steps &  3,000/50,000 \\
    Action size & 100 & Time size & 50 & TCN layers & 6/12\\
    FCN layers & 3/6 & TCN filter & 64 & FCN filter & 128/256 \\
    TCN kernal & 5/7 & TCN dilation & $2^k/3^k$ & Dropout & 0.3\\
    \hline
    \end{tabular}
    \label{tab:param}
\end{table}

\section{Features for Non-neural Models}
We follow our prior work \cite{lou2021predicting} in selecting the summary statistics of activity logs as features for GBM, SVM and RF in Table \ref{tab:prediction}. The features include: a) Workload measures -- total EHR time, after-hours EHR time, patient load, inbox time, time spent on notes, chart review, and number of orders, per patient per day.
b) Temporal statistics -- mean, minimum, maximum, skewness, kurtosis, entropy, total energy, autocorrelation, and slope of time intervals.



\end{appendices}